# Integrated Sensing and Earthmoving Vehicle for Lunar Landing Pad Construction

Volker Nannen[1,2,3], Damian Bover[1,4] and Dieter Zöbel[2,5]

[1]Sedewa, Finca Ecologica Son Duri, C/ Palma-Manacor KM 40, 07250 Vilafranca de Bonany, Spain
[2]University of Koblenz-Landau, Universitätsstraße 1, 56070 Koblenz, Germany
[3]email: vnannen@gmail.com
[4]email: utopusproject@gmail.com
[5]email: zoebel@uni-koblenz.de

**ABSTRACT**
Reducing the forces necessary to construct projects like landing pads and blast walls is possibly one of the major drivers in reducing the costs of establishing lunar settlements. The interlock drive system generates traction by penetrating articulated spikes into the ground and by using the natural strength of the ground for traction. The spikes develop a high pull to weight ratio and promise good mobility in soft, rocky and steep terrain, energy-efficient operation, and their design is relatively simple. By penetrating the ground at regular intervals, the spikes also enable the in-situ measurement of a variety of ground properties, including penetration resistance, temperature, and pH. Here we present a concept for a light lunar bulldozer with interlocking spikes that uses a blade and a ripper to loosen and move soil over short distances, that maps ground properties in situ and that uses this information to construct landing pads and blast walls, and to otherwise interact with the ground in a targeted and efficient manner. Trials on Mediterranean soil have shown that this concept promises to satisfy many of the basic requirements expected of a lunar excavator. To better predict performance in a lunar or Martian environment, experiments on relevant soil simulants are needed.

**INTRODUCTION**

Technical progress and renewed government interest make the establishment of permanent bases on the Moon and Mars look increasingly likely. Because of the high cost of shipping, a key strategy is to bootstrap a lunar or Martian presence by shipping only equipment that is essential to establish an initial presence and manufacture all further equipment through in-situ resource utilization (ISRU). One of the first tasks to accomplish is the provision of a safe landing zone that supports the vertical take-off and landing of rocket-powered vehicles. Metzger et al. (2009) argue that *in situ* construction of landing pads must precede human landing and that we should first construct them on the Moon.

Lunar regolith mostly comprises abrasive fine-grained rock fragments that cover the entire lunar surface in thickness from meters to tens of meters. It also includes larger rock fragments. The density of regolith increases rapidly with depth, reaching 90% at a depth of 30 cm (Carrier 1991; Just 2019). The rocket plume of a vehicle landing on and taking off from the Moon can eject dust grains from the Moon surface which can travel great distances at high velocities, possibly damaging the vehicle and any infrastructure even hundreds of meters away, as no atmosphere slows the grains down (Hintze and Quintana 2013). The spray of loose material also obscures the view of the lander, spoofs and damages sensors, and damages the landing site such that the vehicle may not land safely (Metzger et al. 2009). To mitigate this threat, we need to construct landing pads and blast walls to minimize the number of grains ejected, and to protect the surrounding infrastructure from their impact. While we should seal landing pads with a very hard and robust surface manufactured from local regolith, during the initial phase of the bootstrap such technology will probably not be available. We will have to construct the first landing pads and blast walls with a minimum in equipment weight, operational complexity, and energy so that we can safely land additional resources.

According to Christensen et al. (1968), soil erosion caused by rocket engine exhaust gas is of three basic types: 1. Bearing load cratering: rapid cratering caused by exhaust gas pressure on a soil surface exceeding the bearing capacity of the surface. 2. Viscous erosion: erosion by entrainment of soil particles as the gas flows over the surface. 3. Diffused gas eruption: movement of the soil caused by the upward flow of gas through the pores of the soil during and after engine firing. This suggests that as a first remedy against excessive dust ejection, we remove the loose surface material from the area affected by the landing plume, down to a depth of 30–50 cm. The strength and density of the remaining layer will probably be more resistant to viscous erosion and diffused gas eruption and might even help against bearing load cratering. We can use the excavated material to surround the landing pad with a blast wall. Once the required infrastructure is in place, we can harden the surface with ISRU.

**Bulldozer with ripper.** The earthmoving industry prefers bulldozers to move material over short distances (< 60 m), scrapers for medium distances (60–600 m), and excavators and trucks for longer distances (Derden and Brittain 2014). Since the clearing of landing pads and roads and the construction of blast walls will probably involve the transport of material over short distances, we propose that the first earthmover will be a simple bulldozer, with a blade for grading and pushing attached to the front and a ripper for soil loosening attached to the rear of the device. These tools will need actuators to control the depth or height of engagement and the angle of attack. The operation of such tools requires forces that are mostly parallel to the motion of the vehicle, which simplifies vehicle design and control.

Lunar surface gravity is 1/6 that of Earth, and Martian surface gravity is 38% that of Earth. One problem of working in low gravity environments is that we have a limited understanding of the sloshing and oscillation of a bucket full of granular material in low gravity. Reduced damping and the unusual adhesive and electrostatic properties of lunar granular material make it difficult to predict the kinetics of the material, the bucket, and the vehicle while moving over rough terrain, and how this will affect the workflow and the stability of the vehicle. There is almost no experience from the field, and the problem has received little attention in the literature, making it very difficult to define meaningful margins of safety. Given the state of the art, attaching a bucket to the vehicle, while mechanically feasible, would so much complicate the kinetics of the overall system and put so many additional constraints on its dimensions and weight distribution, that it might not provide any operational advantage over blades and rippers. Another advantage of blades and rippers over buckets is that they significantly increase the maximum slope that a vehicle can operate on, which is relevant for blast wall construction. Rippers and blades can also loosen and move rocks that would not fit into buckets and the requirements on actuators that control a blade or plow are less stringent than for buckets.

**THE INTERLOCK BULLDOZER**

Bulldozers, like all earthmoving equipment, need to develop strong tractive forces to loosen and move the material. Just at al. (2019) provide an overview of requirements for lunar regolith excavation and summarize the current state of the art. Existing solutions generate traction through a frictional connection of wheels or tracks with the ground, which implies that available drawbar pull is a function of vehicle mass. This poses a significant problem in the reduced gravity environment of the Moon, especially for operations that loosen the ground. On Earth, optimal tractive efficiency is achieved at a pull/weight ratio of about 0.4 on most soils, so for every kg of vehicle mass, a draft of 4 N can be developed (Zoz and Grisso 2003). On the Moon, the Apollo Lunar Roving Vehicle could only have achieved a pull/weight ratio of 0.21 in terms of its Lunar weight (Wilkinson 2007). This amounts to 0.35 N of drawbar pull for every kg of vehicle mass at launch, a reduction by an order of magnitude. Reducing the forces necessary for excavation is possibly one of the major drivers in reducing costs associated with establishing lunar settlements (Zacny et al. 2012).

Creager et al. (2012) show that a push-pull device with independently articulated pairs of wheels can double the pull/weight ratio. Ebert and Larochelle (2018) propose dynamic anchors which engage and disengage from the legs of a walking robot to allow for efficient locomotion on soft terrain, but do not address the question of how to penetrate anchors efficiently and reliably into the ground. Bover (2011) discovered that a push-pull device with frames that move independently along a common axis can penetrate interlocking spikes reliably and efficiently into the soil. While the device developed by Creager et al. brakes the set of tires which is pushed backward to provide traction for the other set of non-braked tires, the device developed by Bover has articulated spikes attached to each frame that penetrate the ground when pushed backward until they interlock with the ground and push the other frame forward, extracting the spikes of the other frame in the progress (Nannen et al. 2016; 2017).

Figure 1 shows a current design of the interlock drive system. This device comprises two frames connected by a central axis to which we welded a motorcycle chain. An electric motor attached to the rear frame to the left drives a cogwheel that meshes with the motorcycle chain and pulls the two frames together and pushes them apart in an alternating motion pattern. The rear frame has two undriven wheels for ground clearance and two large double spikes that we attached via lever arms to hinges close to the ground, backward of each frame. The spikes and lever arms are painted in red.

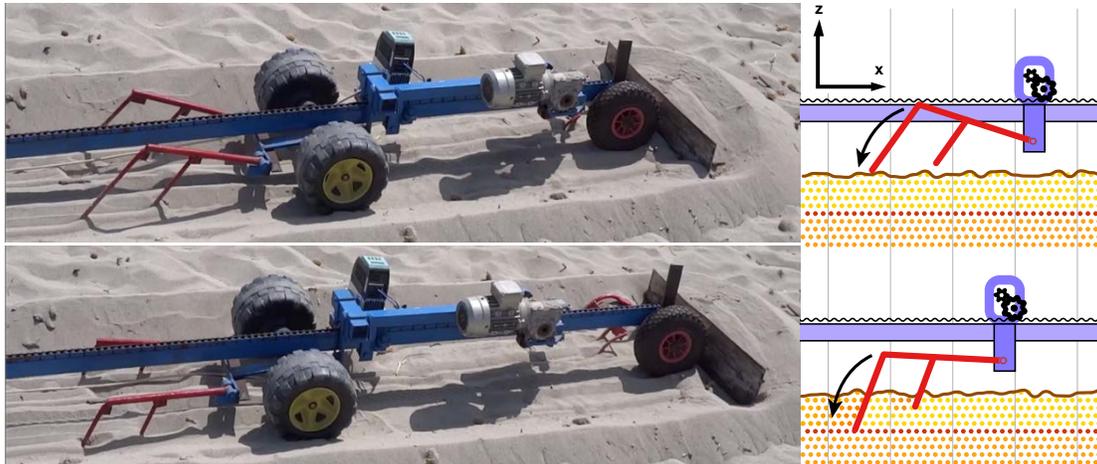

**Figure 1: Push-pull bulldozer with interlocking spikes. Motion is from left to right.**

The larger spikes are made from 16 mm rebar and incline backward and downward at a rake angle $\alpha = 45°$ from horizontal when sliding over the ground and $\alpha = 65°$ when penetrating to a depth of 20 cm. The front frame to the right has a single undriven wheel for ground clearance, a blade to push soil, and small spikes with curved lever arms, half-hidden behind the central axis and painted in red. When the motor moves a frame backward, it penetrates the spikes of that frame into the ground, and when it moves a frame forward, it pulls the corresponding spikes out of the ground.

In the upper photo of Figure 1, the motor has just pulled the rear frame to the left forward and the large spikes are still out of the soil. In the lower photo, the motor has reversed direction and has pushed the rear frame backward until the spikes have penetrated the soil. Once penetration has reached sufficient depth, the front frame to the right and its attached blade push the accumulated soil forward, acting as a bulldozer. For a video of this and other implementations see http://sedewa.com/Xtreme.html.

**Pull/weight ratio.** The direction of the thrust force $F_t$ that pushes an interlocking spike into the ground follows a line from the hinge to the tip of the spike, see Figure 2. Let the thrust angle $\beta$ be the inclination from horizontal of this imaginary line. In the device of Figure 1, we have $\beta = 10°$ when the spike is out of the ground and $\beta = 30°$ at a penetration depth of 20 cm. See Table 1 for a list of symbols.

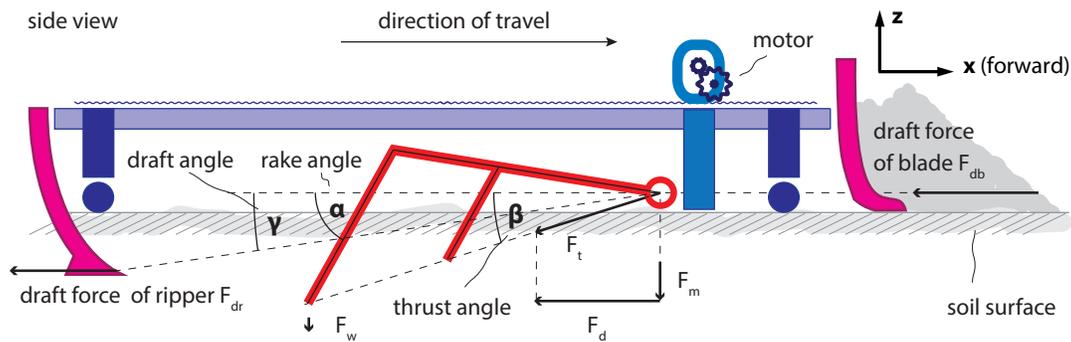

**Figure 2: Forces acting on an interlocking bulldozer.**

**Table 1:** List of symbols

| | |
|---|---|
| α | rake angle, inclination from horizontal of the spike |
| β | thrust angle, inclination from horizontal of the line from hinge to spike tip |
| γ | draft angle, inclination from horizontal of the line from hinge to ripper tip |
| $F_w$ | weight force at the spike tip, about 7 N on Earth in the present configuration |
| $F_d$ | draft force at the hinge |
| $F_{db}$ | draft force of the blade |
| $F_{dr}$ | draft force of the ripper |
| $F_m$ | reactive force of the vehicle mass, $F_m = F_d * \tan \beta$ |
| $F_t$ | the thrust from hinge to the spike tip |

The thrust angle β defines the pull/weight ratio of the vehicle as follows. For a blade, the draft $F_d$ at the hinge is equal to the draft force $F_{db}$ of the blade. It creates a moment force about the tip of the spike, which lifts the vehicle at the hinge with a force of magnitude $F_d \tan \beta$. This lift is countered by a reactive force $F_m$ of equal magnitude from the vehicle mass, provided that lift does not exceed vehicle mass. If the lift does exceed vehicle mass the vehicle will flip over. The device in Figure 1 has β = 30° at the maximum penetration depth of 20 cm, resulting in a pull/weight ratio of $F_d / (F_d \tan 30°) = 1.7$. Even if we double the required vehicle weight for reasons of operational safety, the pull/weight ratio is still 0.85, twice that of a wheeled vehicle.

The center of the draft force $F_{dr}$ of the ripper is below the soil surface, see Figure 2. The line from this center to the hinge has an inclination γ from horizontal, resulting in a dynamic weight transfer from the ripper to the device of magnitude $F_{dr} \tan \gamma$. Hence, the pull/weight ratio for operating the ripper is higher than for operating the blade.

An important property of the interlock drive system is that we can increase the pull/weight ratio by reducing the maximum thrust angle β either by using more spikes (which interferes with steering, see below) or by using a longer lever arm. Doubling the length of the lever arm will halve β and double the pull/weight ratio.

**OPERATIONAL REQUIREMENTS**

Mueller et al. (2008) identify five basic requirements for lunar excavators, which we generalize to include bulldozers: 1) navigate the lunar surface without getting stuck, 2) avoid rocks, 3) traverse 20° slopes when fully loaded, 4) minimize power and peak power, 5) operate reliably over long periods. We discuss these requirements below.

**Mobility, ability to deal with rocks, and ability to climb.** Field trials with several prototypes have shown that the interlock drive system is mechanically robust, reliable and versatile for a range of applications and that it not only does not get stuck but that it generates a strong tractive force on every type of soil we tried, including soft beach sand and soil saturated with water (Nannen et al. 2019). Also, a device with interlocking spikes can climb granular material at an angle of repose of at least 40° (Nannen et al. 2016), see Figure 3. The vehicle in the climbing trials did not carry a load, but easily pushed rocks out of its way that were at least half its weight.

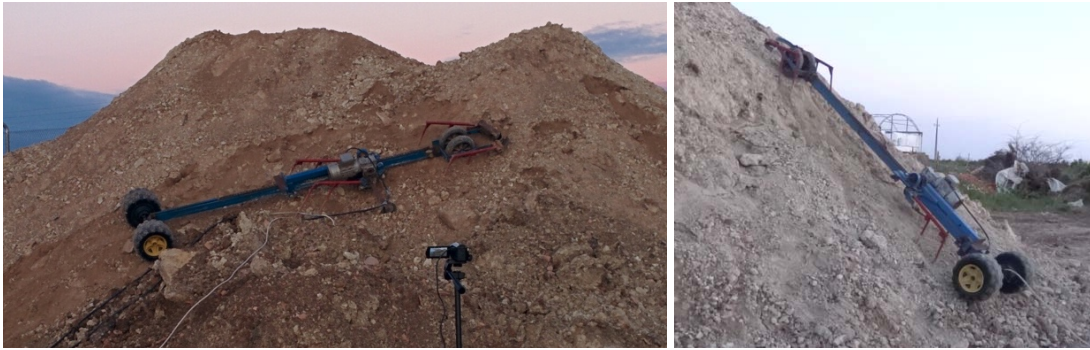

**Figure 3 left: Traversing steep terrain. Right: Climbing an angle of repose of 41°.**

**Power and efficiency.** Once anchored into the ground, a spike does not need energy to provide traction. However, penetrating a spike into the ground requires work. During penetration, the frame that carries the spike slips backward by a certain distance, similar to the slip of a wheel. By multiplying the increments in slip with the draft applied at that moment, we can calculate the cumulative work needed to anchor a spike, i.e., to penetrate a spike deep enough to withstand the applied draft. Figure 5 shows the cumulative work needed to anchor a spike into soil of three different compaction levels (Nannen et al. 2019). We measured the penetration resistance of each compaction level with a cone penetrometer: 960 kPa, 3.9 MPa, and 6 MPa. The x-axis shows the applied draft. The y-axis shows the cumulative work needed to anchor a spike. Figure 5 shows that the cumulative work needed to anchor a spike grows linearly with the applied draft and that the slope depends on the penetration resistance of the soil.

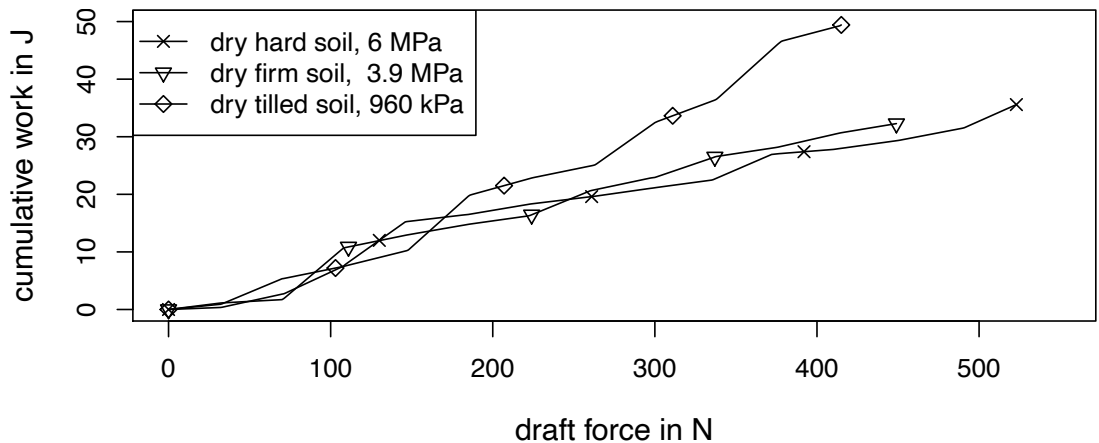

**Figure 5: Work needed to anchor a spike.**

It costs about 30 J to anchor an interlocking spike to withstand 300 N in soft soil. If the spike then pulls this load for two meters, that is 600 J of useful work, at a tractive efficiency of 95%. The spikes must be pulled out and forward during the second part of the motion cycle, which somewhat reduces the overall tractive efficiency. Tractive efficiency increases if the load travels further during each motion cycle. The frames of the device in the right photo of Figure 4 travel four meters along the main axis during each motion cycle. That device also has lever arms that are twice as long as in Figure 3, resulting in a lower thrust angle β and a push/pull ratio twice as high as in Figure 3.

**Operational reliability.** While the spikes worked well on all tested soils, they cannot penetrate rock. They also cannot penetrate duricrust if the spikes have insufficient weight to break into the surface and start penetration. Duricrust is usually formed by the transport of particles and soluble minerals by water. Its presence on Mars has been confirmed and it may be present on the Moon. If we cannot penetrate the surface, and the surface has no other irregularities with which a spike can interlock, we can only achieve traction by friction. To guarantee that the vehicle does not get stuck under such conditions, we need an actuator that can transfer the weight of the device to the spike to either break into duricrust or to generate enough surface friction for traction.

**Steerability.** The present push-pull device has two frames that move rigidly along a common axis. Since the vehicle only moves if it has firmly anchored at least one spike, an anchored spike must be the fixed point of any rotation. If it has anchored only one spike of one frame, its motor pulls the geometric center of friction of the other frame in a straight line toward this spike or pushes it away from it in a straight line. If the anchored spike is not aligned with the central axis, the geometric center of the other frame still moves in a straight line to and from this spike, while the central axis and the orientation of the vehicle rotate around the spike. By alternating the side of the vehicle where it anchors a single spike during contraction and expansion, a push-pull device with interlocking spikes can turn on the spot. Reiser et al. (2019) have shown that as few as three push-pull cycles are needed for a 180° rotation on the spot.

If spikes on either side of the axis penetrate the ground, or a single spike aligned with the central axis penetrates the ground, the vehicle will also turn if the distance between the geometric center of friction of the moving frame and the central axis is greater than the distance between the penetrating spikes and the central axis. This is the case for example in the left photo of Figure 4. The front frame at the bottom of the photo has a single spike that is aligned with the central axis. The brown iron weight to the right of the photo balances the weight of the electric motor to its left. Shifting the weight along the broad blue frame changes the vertical pressure on the tillage tool and the draft on that side of the frame. Moving the weight to the center will bend the path further. Moving it away will first straighten the path and then bend it in the other direction, an effect that can be used for steering and path control.

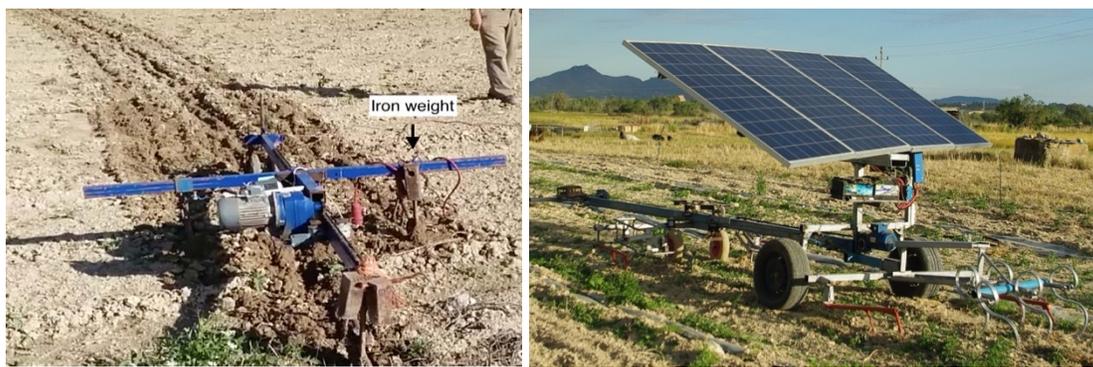

**Figure 4 left: Path control of push-pull tillage device by balancing the draft. Right: Solar powered 1 kW tillage device with interlocking spikes.**

**SENSING CAPABILITY**

Regular penetration of the ground opens up new possibilities for in-situ soil analysis, which may help to create maps of surface strength to identify suitable locations for landing pads and roads, may help to define the depth to which we need to remove the soft surface material during construction, and may help to decide upon the method to harden the surface once we have established the corresponding capabilities for ISRU.

Simple penetration parameters like penetration resistance, motor load, and vehicle motion can provide new types of estimates on soil penetration resistance and shear strength. We found that the force and work needed to anchor an articulated spike do not increase linearly with cone penetration resistance. A cone penetrator is pushed into the soil in a straight line and measures the motion resistance of the most immobile gravel and broken rock along its path. An articulated spike is free to follow the path of least resistance to penetrate a heterogeneous material, allowing the device to estimate a lower bound on the strength of heterogeneous granular material. This lower bound might be relevant for the construction of launch pads, as it may help to localize points of early failure.

As the spikes remain stationary in the soil for several seconds at a time, sensors attached to the spikes can measure temperature, pH, and electric conductivity inside the soil (Bover et al. 2015), all of which might be relevant for the construction of landing pads. Halajian and Reichman (1969) for example find that thermal conductivity correlates with the bearing strength of lunar soil. Figure 6 shows an experimental device with a simple temperature sensor embedded in a cavity at the side of a spike.

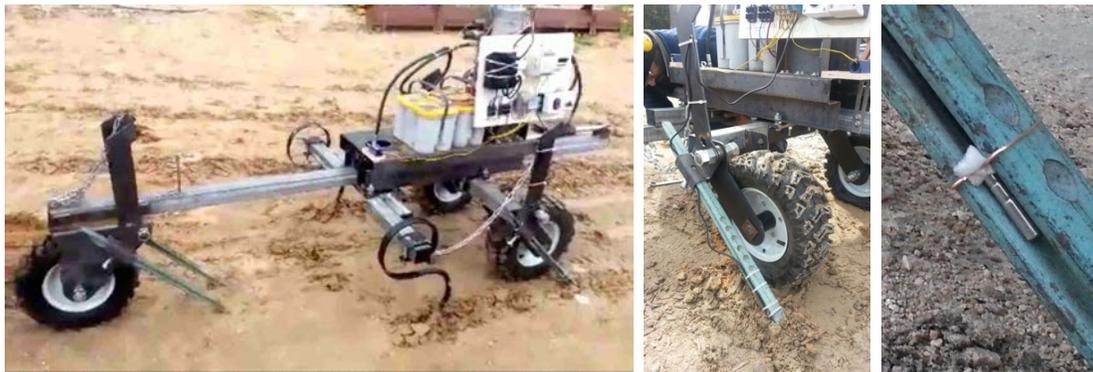

**Figure 6 left: hydraulic-driven push-pull vehicle with interlocking spikes and integrated sensors. Center and right: close up of embedded temperature sensor.**

The field of proximity sensing has led to a host of commercial applications in agriculture, including sensors that penetrate the soil (Adamchuk et al. 2018). Most of these applications require manual operation, a vehicle that regularly stops in the field to penetrate a probe or the collection of samples from a moving vehicle for analysis on the vehicle or in the lab. By contrast, soil penetrating spikes allow unprecedented measurements inside the soil at a high spatial resolution without significant interference with normal vehicle operations.

## CONCLUSION

Landing pads and blast walls are among the first construction projects needed on the Moon. Reducing the forces necessary for excavation is possibly one of the major drivers in reducing the costs associated with establishing lunar settlements. A bulldozer of a given weight can generate significantly more draft with interlocking spikes than with tires or tracks. Interlocking spikes provide mobility in soft and steep terrain, allow for energy-efficient operation, and their design is relatively simple. By penetrating the ground at regular intervals, the spikes also provide an opportunity to measure a variety of ground properties in-situ, including penetration resistance, temperature, and pH. Because they penetrate the soil with force and remain static in the soil for several seconds, they allow detailed mapping of soil parameters that can help to optimize work parameters like the bearing strength of landing pads and roads or select material to construct blast walls. Since all trials reported here were conducted on Mediterranean soil, further experiments on relevant soil simulants are needed to better predict how interlocking spikes would perform in a lunar or Martian environment.

## ACKNOWLEDGMENT

We thank Joseph Cohen for technical advice and the prototype in Figure 6.